\documentclass[graybox]{svmult}
\usepackage{graphicx}
\graphicspath{{./}{Figures/}}
\usepackage{newtxtext}
\usepackage{newtxmath}
\usepackage{booktabs}
\usepackage{tikz}
\usepackage{pgfplots}
\usetikzlibrary{arrows,positioning}
\newcount \double

\definecolor{HighlightColor}{rgb}{0.462745098, 0.725490196, 0.000000000}
\definecolor{HighlightColor2}{rgb}{0.945, 0.349, 0.373}
\definecolor{HighlightColor3}{rgb}{0.349, 0.604, 0.827}
\definecolor{HighlightColor4}{rgb}{0.976, 0.651, 0.353}
\definecolor{HighlightColor5}{rgb}{.62, 0.4, 0.671}

\begin{document}
\title*{Artificial Neural Networks\\generated by Low Discrepancy Sequences}
\author{Alexander Keller and Matthijs Van keirsbilck}
\institute{Alexander Keller \at NVIDIA,
	Fasanenstr. 81,
	D-10623 Berlin, Germany,
	\email{akeller@nvidia.com}
	\and Matthijs Van keirsbilck \at NVIDIA,
	Fasanenstr. 81,
	D-10623 Berlin, Germany,
	\email{matthijsv@nvidia.com}
}
\maketitle

\abstract*{%
.}

\abstract{%
Artificial neural networks can be represented by paths. Generated as
random walks on a dense network graph, we find that the resulting sparse
networks allow for deterministic initialization and even weights with
fixed sign. Such networks can be trained sparse from scratch, avoiding the 
expensive procedure of training a dense network and compressing it afterwards.
Although sparse, weights are accessed as contiguous blocks of memory.
In addition, enumerating the paths using deterministic low discrepancy
sequences, for example the Sobol' sequence, amounts to connecting
the layers of neural units by progressive permutations, which naturally
avoids bank conflicts in parallel computer hardware. We demonstrate
that the artificial neural networks generated by low discrepancy sequences
can achieve an accuracy within reach of their dense counterparts at a much
lower computational complexity.
}

\section{Introduction}

The average human brain has about $10^{11}$ nerve cells, where
each of them may be connected to up to $10^4$ others. Yet, the complexity
of artificial neural networks quite often is determined by fully connected sets
of neurons. Therefore, we investigate algorithms for artificial neural
networks that are linear in the number of neurons and explore
their massively parallel implementation in hardware.

In order to reduce complexity, we motivate the principle of representing an artificial neural
network by paths instead of matrices in Sect.~\ref{Sec:ANN}. Rather than creating such
sparse networks by importance sampling paths from a trained dense network,
training may be much more efficient when considering artificial
neural networks that are sparse from scratch as discussed in
Sect.~\ref{Sec:SparseFromScratch}. Enumerating the paths of an artificial
neural network using a deterministic low discrepancy sequence and
exploiting its structural properties, leads to an efficient hardware
implementation, whose advantages are detailed in Sect.~\ref{Sec:LDgraphs}.
Initial numerical evidence to support the approach is reported
in Sect.~\ref{Sec:Results} before drawing the conclusions. A noteworthy
result is that quasi-Monte Carlo methods enable a completely deterministic
approach to artificial neural networks.

\section{Representing Artificial Neural Networks by Paths} \label{Sec:ANN}

In order to provide an intuition why representing artificial neural networks as
paths may lower their computational complexity, we review their basic principles.

As depicted in Fig.~\ref{Fig:ANN}, the computational graph of a basic artificial
neural network or multi-layer perceptron (MLP) may be organized in $L + 1 \in \mathbb{N}$ layers, each comprising of
$n_l \in \mathbb{N}$ neural units, where $0 \leq l \leq L$. Given an input vector
$a_0$, the output vector $a_L$ is computed layer by layer, where each
vertex determines the activations
\begin{equation} \label{Eqn:ReLU}
  a_{l, i} := \max\Bigg\{0, \underbrace{\sum_{j = 0}^{n_{l-1}-1} w_{l,j,i} \cdot a_{l-1,j}}_{=: z_{l,i}}\Bigg\} .
\end{equation}
For the purpose of the article, it is sufficient to consider
the non-linearity $\max\{0, x\}$ as an activation function,
yielding the so-called rectified linear units (ReLU). The vertices
of the graph are connected by edges with their associated weights $w_{l,j,i} \in \mathbb{R}$.
In summary, each neural unit computes a weighted average of the activations
in the previous layer. If the average is non-positive, it is clipped
to zero, which renders the neural unit inactive. Otherwise the neural unit is called active and
passes on the positive average.

In order to learn the weights from training data, backpropagation~\cite{Backpropagation}
has become the most popular algorithm:
Given an input vector $a_0$ and a desired output vector $d$, the approximation error
\begin{equation} \label{Eqn:BPnorm}
  \delta_L := a_L - d
\end{equation}
is propagated back through the network by computing the
weighted average of the error
\begin{equation} \label{Eqn:BProp}
  \delta_{l-1,i} := \sum_{a_{l,j} > 0} \delta_{l,j} \cdot w_{l,j,i}
\end{equation}
of all active neural units in a layer. If a neural unit is active, the weight
\begin{equation} \label{Eqn:BPupdate}
  w'_{l,j,i} := w_{l,j,i} - \lambda \cdot \delta_{l,j} \cdot a_{l-1, i} \text{ if } a_{l,j} > 0
\end{equation}
of an edge connecting it to
a previous neural unit is updated by the product of the learning
rate $\lambda \in \mathbb{R}^+$, the error at the active neural unit, and the
activation of the previous neural unit.

As formalized by Eqn.~\ref{Eqn:ReLU} and shown in Fig.~\ref{Fig:ANN},
all neural units of one layer are connected to all neural units of the next layer.
Such ``fully connected'' layers are found in many modern artificial neural
networks, for example at the end of classification networks or as so-called
1x1-convolutions, which are fully connected layers with weight sharing across inputs.
Obviously, the computational complexity as well as the number of weights of a layer is
determined by the product of the number of neural units in the current and
previous layer.

\begin{figure}
  \centering
  \begin{tikzpicture}[cnode/.style={draw=gray,minimum width=3mm,circle}]

    \node at (0,-4) {$\vdots$};

    \foreach \x in {0,...,3}
    {   \pgfmathparse{\x+1<4 ? \x : "n_0-1"}
        \node[cnode=blue,label=180:$a_{0,\pgfmathresult}$] (a0-\x) at (0,{-\x-1-div(\x+1,4)}) {};
    }

    \node at (3,-4) {$\vdots$};
    
    \foreach \x in {0,...,3}
    {   \pgfmathparse{\x+1<4 ? \x : "n_1-1"}
        \node[cnode=gray,label={[label distance=0.05cm]270:$a_{1,\pgfmathresult}$}] (a1-\x) at (3,{-\x-1-div(\x+1,4)}) {};

       \foreach \y in {0,...,3}
            \draw [gray] (a0-\y) -- (a1-\x);
    }

    \node at (6,-4) {$\vdots$};

    \foreach \x in {0,...,3}
    {   \pgfmathparse{\x+1<4 ? \x : "n_2-1"}
        \node[cnode=gray,label={[label distance=0.05cm]270:$a_{2,\pgfmathresult}$}] (a2-\x) at (6,{-\x-1-div(\x+1,4)}) {};

       \foreach \y in {0,...,3}
            \draw [gray] (a1-\y) -- (a2-\x);
    }
    
    \node at (9,-4) {$\vdots$};

    \foreach \x in {0,...,3}
    {   \pgfmathparse{\x+1<4 ? \x : "n_L-1"}
        \node[cnode=gray,label=0:$a_{L,\pgfmathresult}$] (aL-\x) at (9,{-\x-1-div(\x+1,4)}) {};

       \foreach \y in {0,...,3}
            \draw [gray] (a2-\y) -- (aL-\x);
    }
    
       \node[ultra thick, cnode=gray, HighlightColor2] (a0-1) at (0,{-0-1-div(0+1,4)}) {};
       \node[ultra thick, cnode=gray, HighlightColor2] (a1-1) at (3,{-1-1-div(1+1,4)}) {};
       \node[ultra thick, cnode=gray, HighlightColor2] (a2-1) at (6,{-0-1-div(0+1,4)}) {};
       \node[ultra thick, cnode=gray, HighlightColor2] (aL-1) at (9,{-3-1-div(3+1,4)}) {};
       \draw [ultra thick, HighlightColor2] (a0-1) -- (a1-1);
       \draw [ultra thick, HighlightColor2] (a1-1) -- (a2-1);
       \draw [ultra thick, HighlightColor2] (a2-1) -- (aL-1);
       \node[ultra thick, cnode=gray, HighlightColor3] (a0-1) at (0,{-2-1-div(2+1,4)}) {};
       \node[ultra thick, cnode=gray, HighlightColor3] (a1-1) at (3,{-1-1-div(1+1,4)}) {};
       \node[ultra thick, cnode=gray, HighlightColor3] (a2-1) at (6,{-2-1-div(2+1,4)}) {};
       \node[ultra thick, cnode=gray, HighlightColor3] (aL-1) at (9,{-0-1-div(0+1,4)}) {};
       \draw [ultra thick, HighlightColor3] (a0-1) -- (a1-1);
       \draw [ultra thick, HighlightColor3] (a1-1) -- (a2-1);
       \draw [ultra thick, HighlightColor3] (a2-1) -- (aL-1);
       \node[ultra thick, cnode=gray, HighlightColor4] (a0-1) at (0,{-3-1-div(3+1,4)}) {};
       \node[ultra thick, cnode=gray, HighlightColor4] (a1-1) at (3,{-2-1-div(2+1,4)}) {};
       \node[ultra thick, cnode=gray, HighlightColor4] (a2-1) at (6,{-1-1-div(1+1,4)}) {};
       \node[ultra thick, cnode=gray, HighlightColor4] (aL-1) at (9,{-2-1-div(2+1,4)}) {};
       \draw [ultra thick, HighlightColor4] (a0-1) -- (a1-1);
       \draw [ultra thick, HighlightColor4] (a1-1) -- (a2-1);
       \draw [ultra thick, HighlightColor4] (a2-1) -- (aL-1);
       \node[ultra thick, cnode=gray, HighlightColor] (a0-1) at (0,{-1-1-div(1+1,4)}) {};
       \node[ultra thick, cnode=gray, HighlightColor] (a1-1) at (3,{-3-1-div(3+1,4)}) {};
       \node[ultra thick, cnode=gray, HighlightColor] (a2-1) at (6,{-2-1-div(2+1,4)}) {};
       \node[ultra thick, cnode=gray, HighlightColor] (aL-1) at (9,{-1-1-div(1+1,4)}) {};
       \draw [ultra thick, HighlightColor] (a0-1) -- (a1-1);
       \draw [ultra thick, HighlightColor] (a1-1) -- (a2-1);
       \draw [ultra thick, HighlightColor] (a2-1) -- (aL-1);
  \end{tikzpicture}
  \caption{Representing the graph of an artificial neural network by paths (colored) instead of
  fully connected layers (including gray) allows for algorithms linear in the number of vertices in time and space.}
  \label{Fig:ANN}
\end{figure}

In order to motivate an algorithm linear in time and space, we rewrite
Eqn.~\ref{Eqn:ReLU}, which is the non-linearity applied to the average,
equivalently as average of the non-linear activation functions:
\[
  z_{l,i} = \sum_{j = 0}^{n_{l-1} - 1} w_{l,j,i} \cdot \max\{0, z_{l-1,j}\}
\]
Considering an integral
\[
  z_{l}(y) := \int_0^1 w_{l}(x,y) \cdot \max\{0, z_{l-1}(x)\} dx
\]
rather than a sum, reveals that layers in artificial neural networks
relate to high-dimensional integro-approximation. Hence, in continuous
form, an artificial neural network is a sequence of linear integral operators
applied to non-linear functions.

From the domain of integral equations, especially the domain of computer graphics,
we know how to deal with such sequences: Sampling path space, we trace light transport
paths that connect the light sources and camera sensors to render synthetic
images. It now is obvious that an
artificial neural network may be represented by paths that
connect inputs and outputs, too.
Computation only along the paths (colored in Fig.~\ref{Fig:ANN})
results in a complexity in space and time that is
linear in the number of paths times the depth of the neural networks
and hence may be linear in the number of neural units.

\begin{figure}
  \centering
  \begin{tikzpicture}
	\begin{axis}[
		legend cell align=left,
      		xlabel=Fraction of connections sampled from fully connected layers, legend pos=south east, legend style={draw=none},
		ylabel=Test Accuracy, xmin=0, xmax=1, ymin=0, ymax=1, height=5.5cm, width=\linewidth]
	\addplot[thick,color=HighlightColor2,mark=*] coordinates {
		(.1, .9670)
		(.2, .9764)
		(.3, .9780)
		(.4, .9804)
		(1, .9784)
	};
	\addlegendentry{LeNet on MNIST}
	\addplot[thick,color=HighlightColor3,mark=*] coordinates {
		(1,		.7311)
		(.90,	.7441)
		(.80,	.7380)
		(.70,	.7358)
		(.60,	.7447)
		(.50,	.7388)
		(.40,	.7385)
		(.30,	.7486)
		(.20,	.7454)
		(.10,	.7380)
		(.09,	.7311)
		(.08,	.7369)
		(.07,	.7303)
		(.06,	.7273)
		(.05,	.7330)
		(.04,	.7204)
		(.03,	.6984)
		(.02,	.6973)
		(.01,	.591)
	};
	\addlegendentry{LeNet on CIFAR-10}
	\addplot[thick,color=HighlightColor,mark=*] coordinates {
		(1,		.7436)
		(.90,	.7470)
		(.80,	.7303)
		(.70,	.7404)
		(.60,	.7435)
		(.50,	.7317)
		(.40,	.7467)
		(.30,	.7555)
		(.20,	.7310)
		(.10,	.7748)
		(.09,	.7715)
		(.08,	.7452)
		(.07,	.7572)
		(.06,	.7624)
		(.05,	.7696)
		(.04,	.7645)
		(.03,	.7642)
		(.02,	.7662)
		(.01,	.7496)
		(.009,	.7596)
		(.007,	.7578)
		(.005,	.7655)
		(.003,	.7676)
		(.001,	.7081)
		(.0005,	.0454)
	};
	\addlegendentry{AlexNet on CIFAR-10}
	\addplot[thick,color=HighlightColor4,mark=*] coordinates {
		(0,	0)
		(0.05,	.1488095188)
		(0.1,	.8035713959)
		(0.125,	.7916666412)
		(0.185,	.794642868)
		(0.25,	.8005952454)
		(0.5,	.8035713959)
		(1,	.815476150)
	};
	\addlegendentry{Top-5 accuracy AlexNet on ILSVRC12}
	\addplot[thick,color=HighlightColor5,mark=*] coordinates {
		(0.05,	0)
		(0.10,	.625)
		(0.13,	.616071434)
		(0.19,	.6011904526)
		(0.25,	.6190476227)
		(0.50,	.5982142639)
		(1.00,	.5892856979)
	};
	\addlegendentry{Top-1 accuracy AlexNet on ILSVRC12}
	\end{axis}
  \end{tikzpicture}
  \caption{Test accuracy of a selection of classic
artificial neural networks for image recognition. Using only about 10\% (or even less) of the connections
of the original trained networks does not result in a notable loss in
test accuracy, indicating potential efficiency gains. The paths through the fully connected layers of the
artificial neural networks have been sampled proportionally to the trained weights.}
  \label{Fig:ANNsparse}
\end{figure}

\subsection{Quantization of Artificial Neural Networks by Sampling}

In \cite{InstantANNQuantization} we derived an algorithm that quantizes a
trained artificial neural network such that the resulting complexity may be
linear. To create these paths given a trained neural network, we exploit an
invariant of the rectified linear units (ReLU): In fact, scaling the activations by a
positive factor $f \in \mathbb{R}^+$ and dividing
the weighted average by the same factor leaves the result unchanged.

Choosing this factor as the one-norm of the weights of a neural unit,
the factor can be propagated forward through the neural network,
leaving the weights of each neural unit as a discrete probability density.

Given the $n$ weights of a single neural unit, assuming $\sum_{k=0}^{n-1}|w_k| = \|w\|_1 = 1$ and defining a
partition of the unit interval by $P_m := \sum_{k=1}^{m}|w_k|$,
it is straightforward to trace paths from the outputs back to the inputs by
sampling proportional to the discrete densities.
The graphs in Fig.~\ref{Fig:ANNsparse} provide evidence
that sampling only a fraction of the connections results in no notable
degradation in the test accuracy. A similar approach can be used to quantize not
just weights, but also activations and gradients to arbitrary precision. For
more details and data, we refer to \cite{InstantANNQuantization} and
\cite{GradientQuantization}.

Above we used the $L^1$-norm to generate probabilities from the weights. It is also
possible to use more advanced importance estimation techniques. This can be used
both during training to precondition the model so it can be pruned to higher
sparsity levels \cite{molchanov2019importance}. Similarly, subsets of neurons
or weights can be selected such that 
a certain sparsity level is maintained throughout training for increased
efficiency \cite{dettmers2019sparsefromscratch,jayakumar2020top}.

\subsection{Sampling Paths in Convolutional Neural Networks (CNNs)} \label{Sec:CNN}

Convolutional neural networks \cite{LecunCNN} contain layers that compute features
by convolutions. For example, common features include first and second derivatives to identify
edges of different orientations in an image.

A convolutional neuron (also called filter or kernel) is specified by a 3D tensor
of weights of width $w$, height $h$, and depth $c_{\text{in}}$ (also called channel dimension). 
Typically, the dimensions $w \times h$ are small, for example, $3 \times 3$.
Given an input tensor of width $W$, height $H$, and depth $c_\text{in}$, each 2D depth slice
is convolved with the corresponding 2D depth slice of shape $w \times h$ of the
weight tensor. Then the resulting features
are summed along the depth dimension to produce one output feature channel of shape $W \times H \times 1$.
With $c_{\text{out}}$ convolutional neurons each computing one output channel from the input tensor, 
a CNN layer may be interpreted as a function that maps $c_\text{in}$ input channels to $c_\text{out}$ output channels, just like an MLP.

Hence, to create a sparse CNN, we trace an edge of a path the same way as for MLPs: by selecting one of the $c_{\text{in}}$ input channels,
and one of the $c_{\text{out}}$ convolutional neurons in the layer.
This activated edge means the selected $w \times h$ depth slice in the 3D weight tensor of the selected neuron
will be convolved with the corresponding slice of the input tensor.

Many recent CNN architectures such as MobileNet, DenseNet, or
QuartzNet \cite{QuartzNet} use $1 \times 1$-convolutions, where $w = h = 1$.
This important special case amounts to a structure identical to a fully connected layer \cite{huang2017densely,changpinyo2017power}, 
where weights are reused across the elements of the input tensor.
Paths are traced in the same way as described before.

Tracing paths through trained convolutional networks is related to ``channel pruning'' \cite{molchanov2019importance}.
This enables coarse sparsity on the filter level, which is more efficient on
current hardware than fine-grained sparsity \cite{changpinyo2017power}, i.e. selecting single weights.

\begin{figure}
  \centering
  \begin{programcode}{Inference with an Artificial Neural Network represented by Paths}
\begin{verbatim}
// initialization

int neurons = 0;

for (int l = 0; l < layers; ++l)
{
  for (int p = 0; p < paths; ++p)
    index[l][p] = neurons + (int) (drand48() * neuronsPerLayer[l]);
  
  neurons += neuronsPerLayer[l];
}

for (int l = 1; l < layers; ++l)
  for (int p = 0; p < paths; ++p)
    weight[l][p] = initialWeight; // deterministic instead of random

float *a = new float[neurons];
float *error = new float[neurons];

// train by backpropagation

...

// inference

for (int i = 0; i < neuronsPerLayer[0]; ++i)
  a[i] = inputs[i];

for (int i = neuronsPerLayer[0]; i < neurons; ++i)
  a[i] = 0.0f; // or bias[i], if bias terms are used

for (int l = 1; l < layers; ++l)
  for (int p = 0; p < paths; ++p)
    if (a[index[l – 1][p]] > 0.0f)  // ReLU
       a[index[l][p]] += weight[l][p] * a[index[l – 1, p]];
\end{verbatim}
  \end{programcode}
  \caption{Implementation of an artificial neural network
  represented by paths using rectified linear units (ReLU) as
  activation functions. Given the numbers of \texttt{layers},
  \texttt{paths}, and the array of \texttt{neuronsPerLayer}, the
  array \texttt{index} stores the indices of neural units along
  a path \texttt{p} created by randomly selecting a neural unit per layer.
  Before training by backpropagation, the
  \texttt{weight} of each edge is set to a constant \texttt{initialWeight}.
  For inference, the activations \texttt{a} of the first layer are set to the
  \texttt{input} data, while the remaining activations are set to zero.
  Enumerating all activations for all subsequent layers and for all
  paths, each activation along an edge is updated, if its previous
  activation along the path is active, i.e. larger than zero, which
  amounts to the rectified linear unit (ReLU) activation function.}
  \label{Fig:Code}
\end{figure}

\section{Training an Artificial  Neural Network Sparse from Scratch} \label{Sec:SparseFromScratch}

Quantizing an artificial neural network by sampling paths still requires to train
the full network. This complexity issue may be resolved by training
an artificial neural network sparse from scratch as principled by
the implementation provided in Fig.~\ref{Fig:Code}.
We start by storing random paths as an array \texttt{index[][]} of indices and enumerate all layers and paths. In layer $l$ of path $p$,
the index of a neural unit is just randomly selected among the
neuron units of that respective layer.

The evaluation - also called inference - of an artificial neural network represented by paths
first copies all inputs to an array of activations and then
initializes all other activations to zero.
The actual computation then loops over all layers and paths,
where an activation is updated
only if the previous vertex along the path $p$ is positive, meaning active.
This is an implicit implementation of the rectified linear neural unit (ReLU)
introduced in equation~\ref{Eqn:ReLU}.

In the same manner, restricting training by backpropagation~\cite{Backpropagation}
as defined by equations~\ref{Eqn:BPnorm}, \ref{Eqn:BProp}, and \ref{Eqn:BPupdate}
to the representation by paths is straightforward.
In analogy to Sect.~\ref{Sec:CNN}, convolutional layers can be represented
by paths and be trained sparse from scratch, too, resembling methods to create
predefined sparse kernels \cite{pSConv,kundu2020predefined}.

It is obvious that the complexity of inference and training is linear in the
number of paths times the depth of the neural network. Also note,
that although sparse, all weights are accessed in linear order, which is the most
efficient memory access pattern on modern computer hardware.

For random paths, multiple paths may select the same weights while leaving others untouched.
This wastes memory and computation and may make training more difficult.
A solution to this issue is to use Low Discrepancy Sequences to generate the paths,
as will be discussed in Sect.~\ref{Sec:LDgraphs}.

\subsection{Constant Initialization} \label{Sec:Init}

For a fully connected neural network, all neurons in a layer have the same connectivity,
i.e. each neuron is connected to the same neurons in the previous and
following layer. If all weights were initialized uniformly, all neurons would receive the same
updates, and the network would learn nothing during training. The usual way to prevent this is
to initialize the weights by sampling randomly from some distribution. 
There has been a lot of work done on finding good initializations, depending on
the used activation function, size of weight tensor, and other factors
\cite{glorot2011deep,he2015delving,de2016overview}.

However with sparse networks, each neuron has a different connectivity pattern.
Instead of introducing randomness by sampling random weights, the non-uniform connectivity pattern
ensures that not every neuron learns the same thing.
This allows one to get rid of the random initialization, as shown in the code in
Fig.~\ref{Fig:Code}, where the weights along the edges of the paths
are initialized with a constant. 

The value of the constant itself is still important as it controls the operator
norm of the affine transformation that each neuron performs. Following the
analysis of \cite{he2015delving}, and considering that our networks use the ReLU
activation function, we use $w_{\text{init}} = \frac{6}{\sqrt{\texttt{fan\_in} + \texttt{fan\_out}}}$,
where \texttt{fan\_in} is the number of inputs to a neural unit and \texttt{fan\_out}
the number of its connections to the next layer.
Biases are initialized with 0, and scale and shift parameters of batch
normalization layers are initialized with 1 and 0, respectively.

We may conclude that the classic random weight initialization required to
make fully connected layers learn is replaced by the fact that
in a artificial neural network sparse from scratch
neural units don't share the same set of connections.

\subsection{Non-negative Weights} \label{Sec:NonNegative}

Weights may change sign during training. Yet, it has been observed
that a graph of an artificial neural network including static signs of the
weights may be separated from training the magnitudes of the weights
\cite{LotteryTicket,DeconstructingLotteryTicket}. However, finding this
graph and its associated static signs requires pruning a trained, fully connected,
and randomly initialized artificial neural network.

When working with optical implementations of artificial neural networks,
only non-negative weights may be used, because either there is light or not \cite{Farhat:85}.
The lack of negative weights is accounted for by amplifying differences of
weighted sums similar to how operational amplifiers work.
A modern example are ternary quantized artificial neural networks
\cite{Ternary2}, where a first binary matrix accounts for all
non-negative weights of a layer, and a second binary matrix produces
the sums to be subtracted. Still, finding the ternary representation
requires pruning a fully connected network and retraining.

Representing artificial neural networks by paths, we propose to attach one fixed
sign to each path. As paths are generated by random walks, selecting the
weights of the even paths to be non-negative and the weights of the odd paths to be
non-positive perfectly balances the number of positive and negative weights.
Alternatively, any ratio of positive and negative weights may be realized
by for example determining the sign of a path by comparing its index to
the desired number of positive paths (even per layer). This architecture
can be thought of as an inhibiting network superimposed on a
supporting network. Such a network may be trained by, for example,
backpropagation with the only restriction that weights cannot
become negative.

\subsection{Normalization} \label{Sec:Normalize}

Using a constant value for weight initialization
allows one to fulfill normalization constraints. For example,
knowing the number of edges incident to a neural unit, it
is straightforward to determine the initial weight such that any
selected $p$-norm of a set of weights will be one. Uniformly
scaling the initial weights allows one to control the operator
norm of the artificial neural network represented by paths \cite{InstantANNQuantization}.

\section{Low Discrepancy Sequences to enumerate Network Graphs} \label{Sec:LDgraphs}

Low discrepancy sequences \cite{Nie:92,NetsSequences}
may be considered the deterministic counterpart of pseudo-random number generators. 
Abandoning the simulation of independence, they
generate points in the unit hypercube much more uniformly distributed
than random numbers ever can be. Improving convergence speed, they have become
the industry standard for generating light transport paths
in computer graphics \cite{NutshellQMC,ScreenSpaceBlueNoise}.

Reviewing classic concepts of parallel computation (see Sect.~\ref{Sec:Parallel}),
taking advantage of the properties of low discrepancy sequences
(see Sect.~\ref{Sec:LD}) yields an algorithm to enumerate
the graph of an artificial neural network (see Sect.~\ref{Sec:QRN})
that perfectly suits an implementation in massively parallel hardware (see Sect.~\ref{Sec:CollisionFree}).

\begin{figure}
  \centering
  \includegraphics[width=0.3\linewidth]{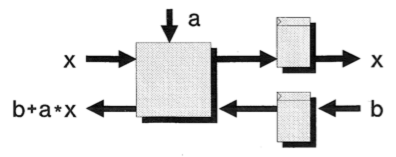} \hfill
  \includegraphics[width=0.68\linewidth]{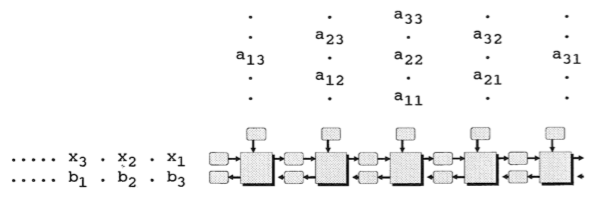}

\bigskip
\begin{tikzpicture}
\foreach \x in {0,1,..., 15}
\node (top\x) at (0.4*\x, 1)[rectangle,draw=black][scale=1] {};
\foreach \x in {0,1,..., 15}
\node (bottom\x) at (0.4*\x, 0)[rectangle,draw=black][scale=1] {};
\foreach \x in {0,1,..., 7}{
\double=\x
\multiply\double by 2
\draw[->, -latex] (bottom\x.north)--(top\the\double.south);
}
\foreach \x in {8,9,..., 15}{
\double=\x
\advance\double by -8
\multiply\double by 2
\advance\double by 1
\draw[->, -latex] (bottom\x.north)--(top\the\double.south);
}
\end{tikzpicture}

  \caption{Parallel computer architecture. Top: Systolic arrays tile identical
  processing units for parallel processing. Here, instances of a multiply-and-add
  unit with input and output registers are chained to parallelize matrix multiplications.
  Bottom: Linking registers and processing units using the interleaving permutation
  as given by a perfect shuffle allows for parallelizing many useful computations, the
  most prominent example being the fast Fourier transform (FFT).}
  \label{Fig:Parallel}
\end{figure}

\subsection{Parallel Computer Architecture} \label{Sec:Parallel}

Already in the 1970s, the concepts of systolic arrays \cite{SystolicArrays}
and the perfect shuffle \cite{PerfectShuffle} have been investigated
in the context of parallel computer architecture.

Systolic arrays are based on simple processing units that are
chained to form pipelines.
As an example, the top of Fig.~\ref{Fig:Parallel} shows a multiply-and-add-unit that
computes $a \cdot x + b$. Both $x$ and $b$ are buffered by a register.
Chaining multiple such compute units allows one to parallelize large
parts of matrix multiplication. Obviously the latency of a systolic
array pipeline is determined by the length of the chain of processing
elements. The notion ``systolic'' stems from the analogy to the heart pumping
in and out blood within a heartbeat, where in systolic arrays data is
pumped in and out within a cycle. In fact,
Google's tensor processing units (TPU) are based on this architecture.

Perfect shuffle networks connect an array of registers to an array of
processing units by the permutation resulting from perfectly
interleaving two decks of cards one by one from each deck
(see the bottom part of Fig.~\ref{Fig:Parallel}). The architecture
has many famous applications, including the efficient
implementation of the fast Fourier transform (FFT).
The number of iterations is the latency, which amounts to
the logarithm in base 2 of the number
of registers, i.e. values to process.

Unrolling the iteration results in a structure reminiscent of the
layer structure of artificial neural networks (see Fig.~\ref{Fig:ANN}) and in fact has been
tried to construct simple optical neural networks \cite{PerfectShuffleANN}.Yet, 
the connection pattern of the perfect shuffle appears to be too restrictive.
In a series of articles~\cite{dey2017interleaver,SparseNNHardware,dey2018characterizing}
more general permutations to connect layers have been explored.
The permutations have their origins in interleaver design and interleaved codes.
Visualizing the connection patterns in the unit square \cite[Figs.~4 and 5]{dey2017interleaver},
where a point means a connection
of the neuron at coordinate $x$ to a neuron in the next layer at coordinate $y$
in the subsequent layer, tends to make one think of sampling patterns
as used in random number generation and quasi-Monte Carlo methods \cite{Nie:92}.

\subsection{Progressive Permutations} \label{Sec:LD}

Many low discrepancy sequences are based on radical inversion. The
principle is  best explained by taking a look at the
van der Corput sequence
\begin{eqnarray*}
  \Phi_b: \mathbb{N}_0 & \rightarrow & \mathbb{Q} \cap [0,1) \nonumber \\
   i = \sum_{l = 0}^\infty a_l(i)  b^l & \mapsto & \Phi_b(i)
    := \sum_{l = 0}^\infty a_l(i)  b^{- l - 1}
\end{eqnarray*}
in base $b \in \mathbb{N} \setminus \{1\}$ that maps the integers to the unit interval: Representing the integer
$i$ as digits $a_l(i)$ in base $b$ and mirroring this representation at the decimal point
yields a fraction between zero and one. 

For contiguous blocks of indices $k \cdot b^m \leq i < (k + 1) \cdot b^m - 1$ for any
$k \in \mathbb{N}_0$, the radical inverses $\Phi_b(i)$ are equidistantly
spaced. As a consequence of this perfect stratification, the integers
$\lfloor b^m \Phi_b(i) \rfloor$ are a permutation of $\{0, \ldots, b^m - 1\}$.
Fixing $b^m$, $k$ enumerates a sequence of permutations.
As an example for $b = 2$, the first $2^4 = 16$ points
yield the permutation
\[
  16 \cdot \left. \Phi_2(i) \right|_{i=0}^{15} = (0,8,4,12,2,10,6,14,1,9,5,13,3,11,7,15) .
\]

These properties are shared by the individual components of the
$s$-dimensional sequence \cite{Sobol:67}, which may be the most popular
low discrepancy sequence: Its first component is $x^{(0)}_i := \Phi_2(i)$, while
the subsequent components
\begin{equation} \label{Eqn:Sobol}
  x_i^{(j)} = (2^{-1} \cdots 2^{-m}) \cdot
  \underbrace{\left(C_j \cdot \left(\begin{array}{c}a_0(i) \\ \vdots \\ a_{m-1}(i) \end{array}\right)\right)}_\text{in $\mathbb{F}_2$}
  \in \mathbb{Q} \cap [0,1)
\end{equation}
multiply a generator matrix $C_j$ with the vector of digits before radical inversion.
The generator matrices $C_j$ are determined by the $j$-th primitive polynomial
and for more details we refer to \cite{Sobol:67,JK03,SobolJK,NetsSequences}.

The matrix vector multiplication takes place in the field $\mathbb{F}_2$ of two
elements and very efficiently can be implemented using bit-wise parallel operations
on unsigned integers. For each digit set in the integer $i$, the corresponding column of the 
generator vector just needs to be xor-ed with the so far accumulated value:

{\small
\begin{verbatim}
unsigned int x = 0;

for (unsigned int k = 0; i; i >>= 1, ++k)
  if (i & 1)
       x ^= C[k]; // parallel addition of column k of the matrix C
\end{verbatim}}

Experimenting with the Sobol' sequence \cite{Sobol:67} is very practical,
because an efficient implementation \cite{JK03} along with the source code and
generator matrices has been provided at \url{https://web.maths.unsw.edu.au/~fkuo/sobol/}.
Blocking groups of bits during radical inversion allows for an even faster generation
of the Sobol' sequence, see~\cite[Listing~3.2]{CarstenPhD}.

The permutation properties described above are a consequence of each
component of the Sobol' sequence being a $(0,1)$-sequence in base
$b = 2$. As the Sobol' sequence produces Latin hypercube samples
for each number of points being a power of 2, it can be used to
create permutations in a progressive way. For
more detail on $(t,s)$-sequences that in fact are sequences of
$(t,m,s)$-nets, we refer to \cite[Ch.~4]{Nie:92}.

\begin{figure}
  \centering
  \includegraphics[width=0.49\linewidth]{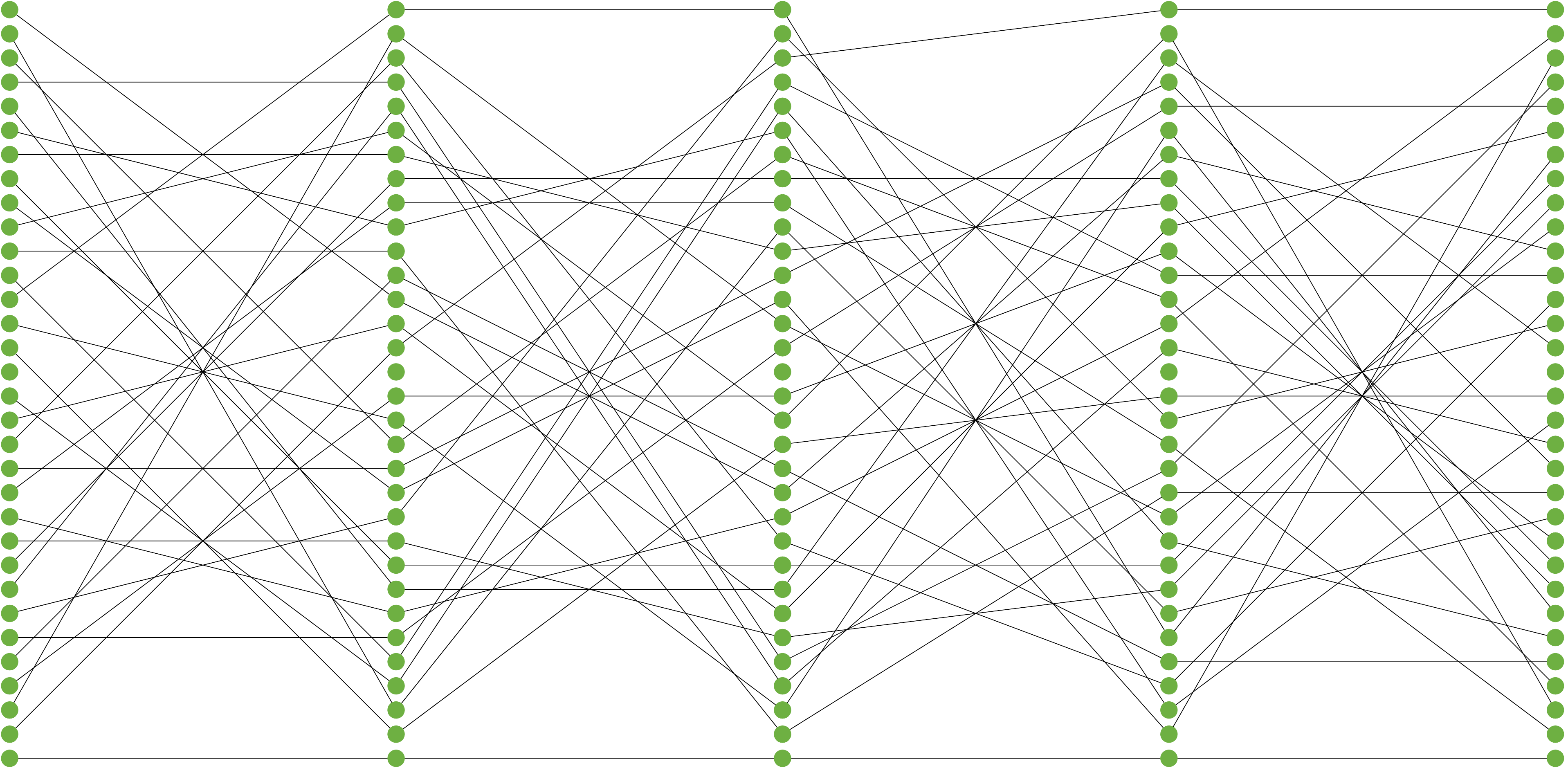}
  \hfill
  \includegraphics[width=0.49\linewidth]{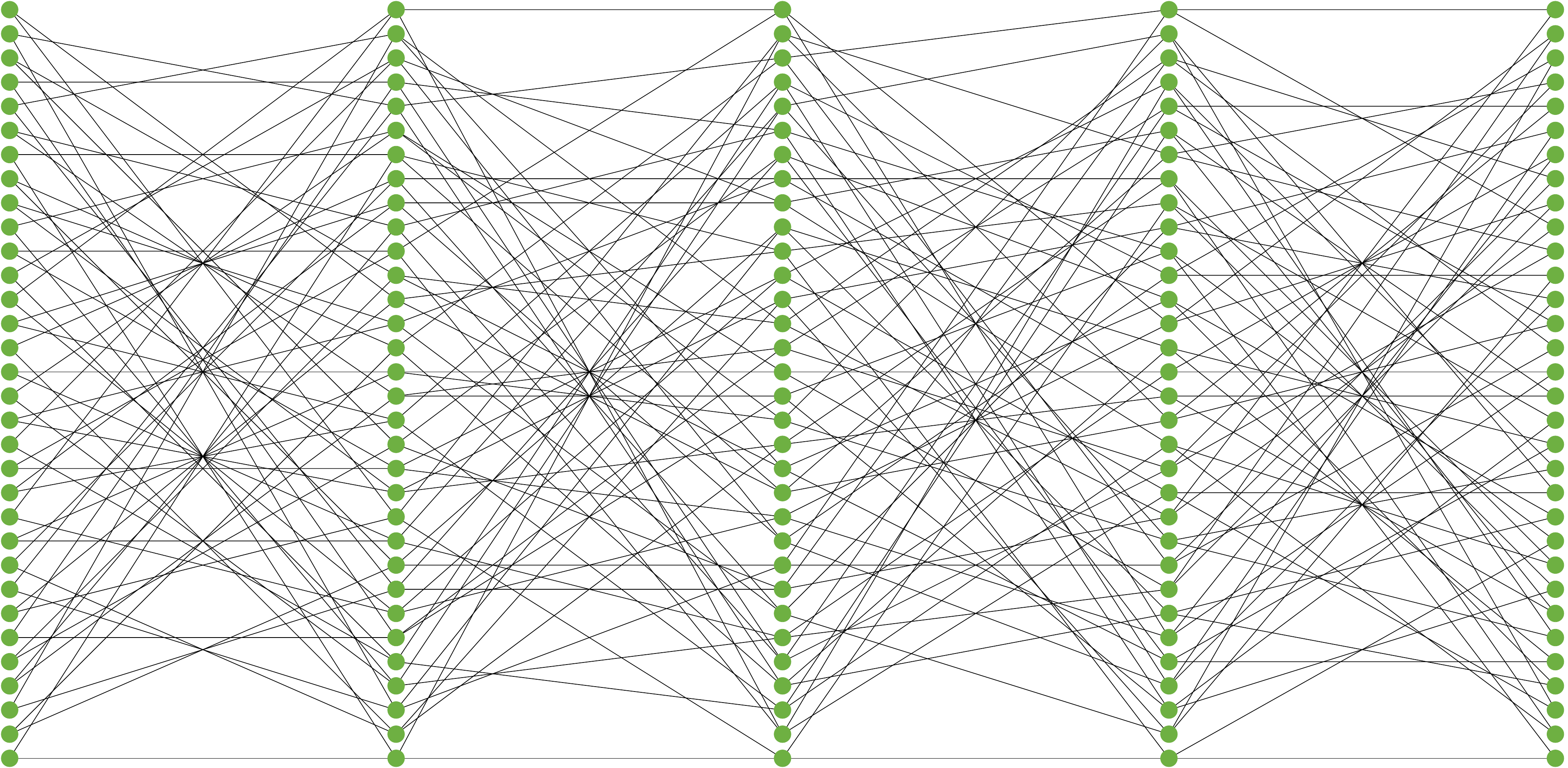}
  
  \bigskip
  
  \includegraphics[width=0.49\linewidth]{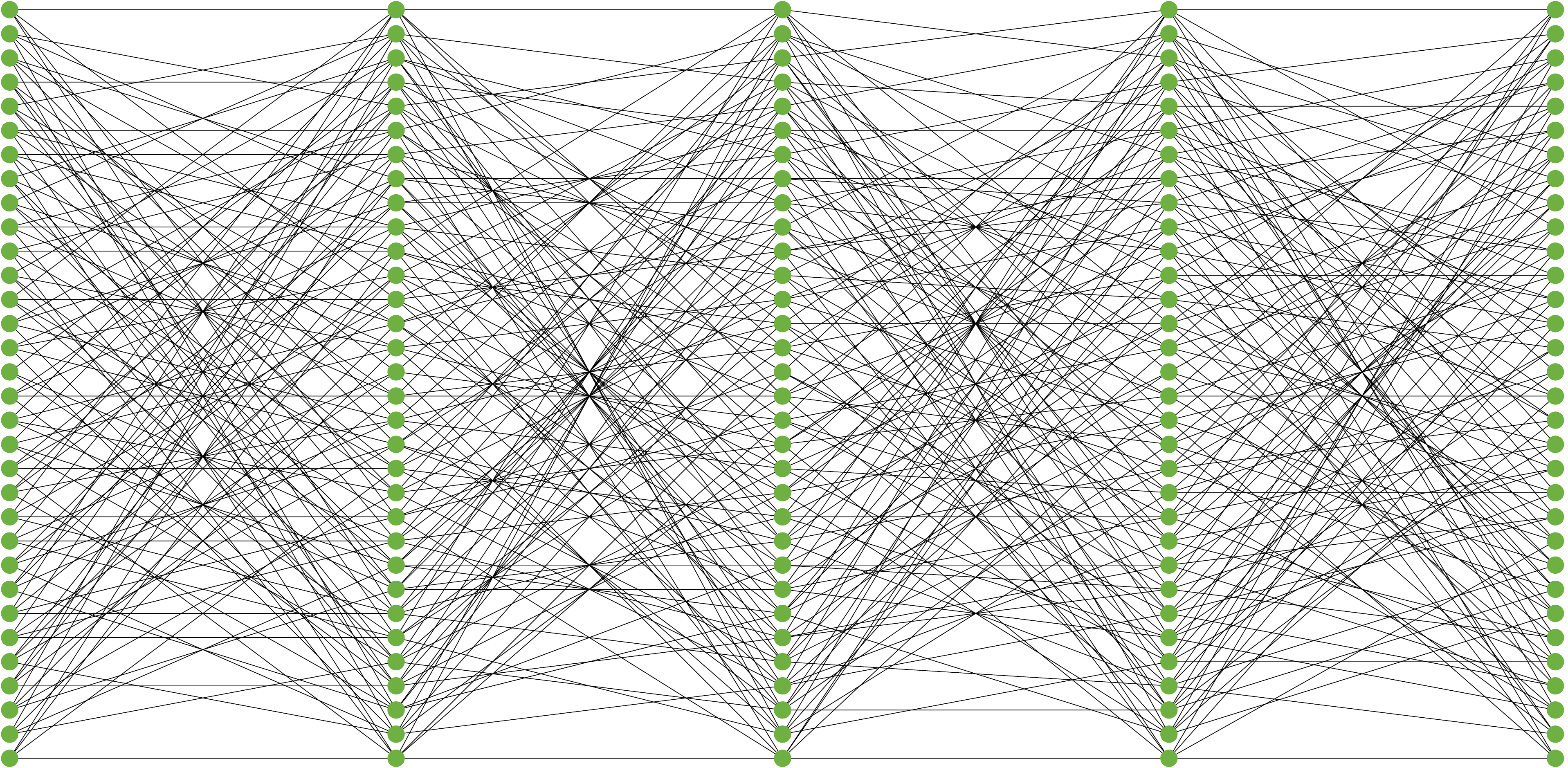}
  
  \caption{Progressive enumeration of paths: For each 32 neural units
  in 5 layers, 32 (top left), 64 (top right), and 128 (bottom) paths generated
  by the Sobol' sequence are shown. The
  number of paths per neural unit  is 1, 2, or 4, respectively.}
  \label{Fig:ANNprogressive}
\end{figure}

\begin{figure}
  \centering
  \includegraphics[width=0.49\linewidth]{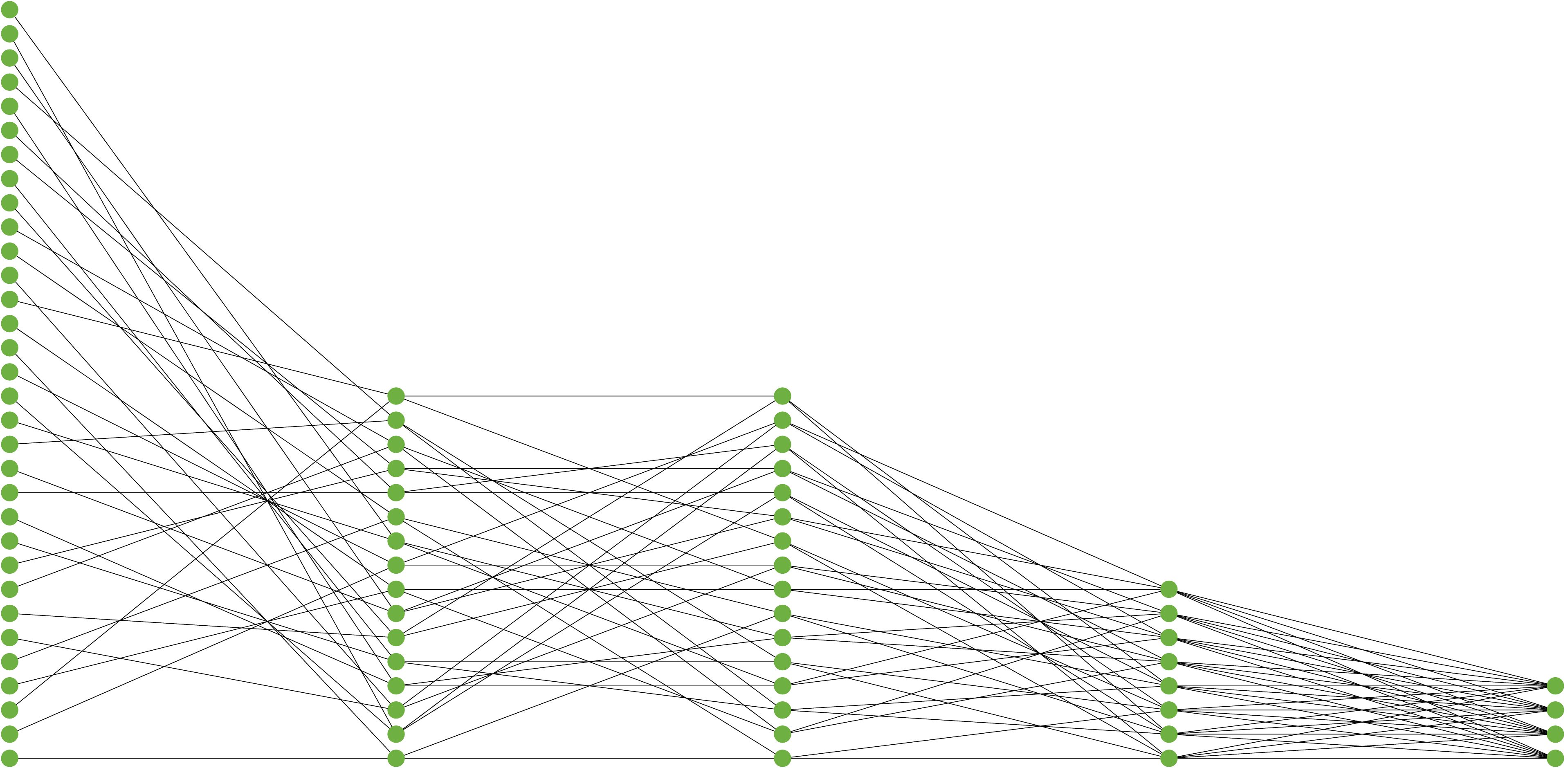}
  \hfill
  \includegraphics[width=0.49\linewidth]{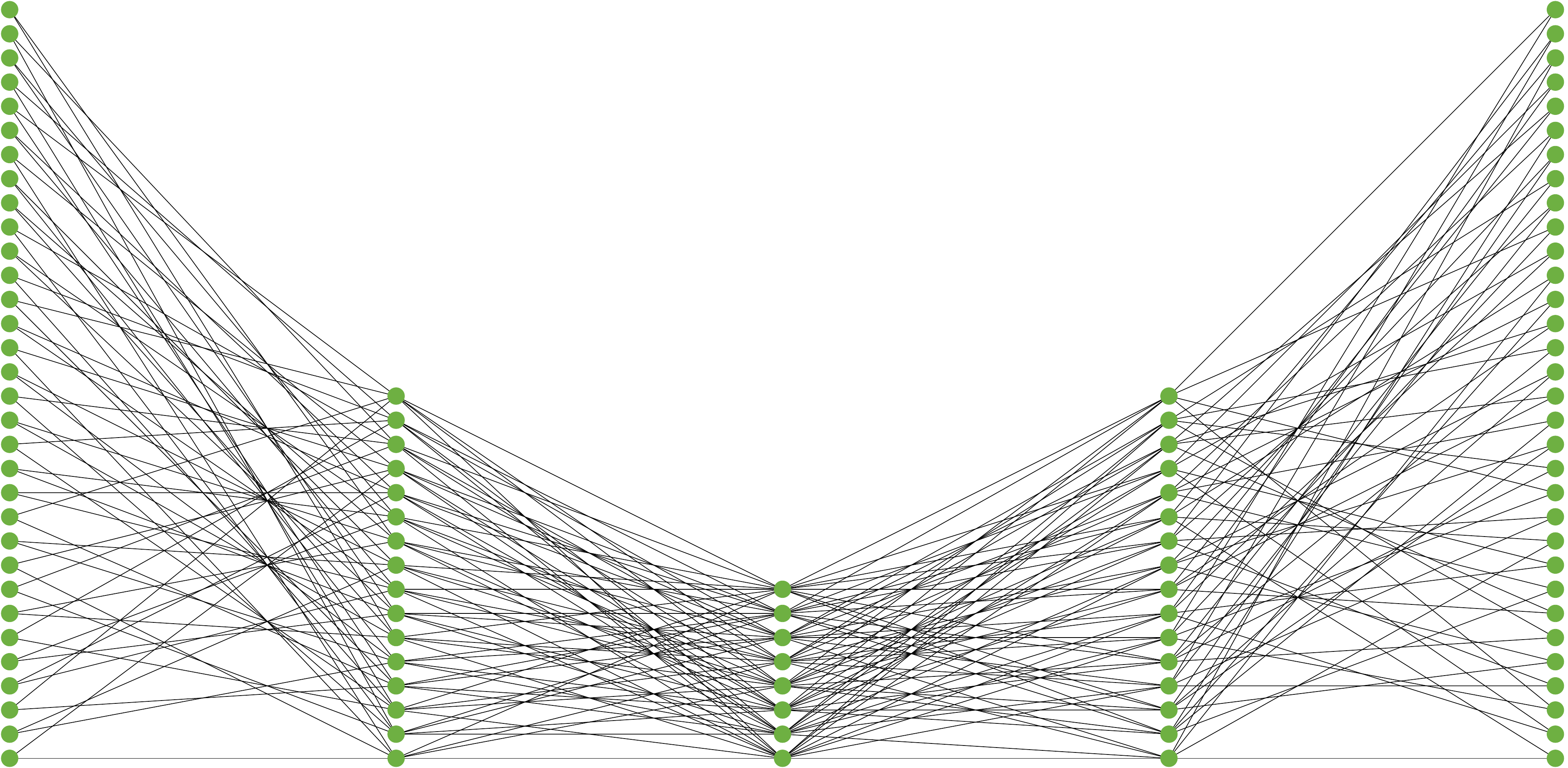}
  
  \caption{Illustration of classic network architectures generated by the
  Sobol' sequence. Left: 32 inputs are encoded to 4 outputs. Such architectures
  are typically used for classification tasks. Right: 32 inputs are encoded to a
  latent space of 8 neural units and decoded back to 32 outputs. This is a common
  architecture of auto-encoders, whose typical task is filtering signals. Note that
  the number of neural units in each layer and the number of paths are powers
  of 2 and the fan-in and fan-out is constant across each layer.}
  \label{Fig:ANNarchitectures}
\end{figure}

\subsection{Sampling Quasi-Random Paths} \label{Sec:QRN}

We will use the components of the Sobol' sequence instead of the pseudo-random
number generator that sampled the path indices of the sparse networks in Sect.~\ref{Sec:SparseFromScratch}.
As contiguous blocks of lengths of powers of 2 form permutations,
we choose a power of 2 neurons per layer. This links the neural units
\begin{equation} \label{Eqn:Link}
  \left( a_{l, \left\lfloor n_l \cdot x_i^{(l)} \right\rfloor}, a_{l+1, \left\lfloor n_{l+1} \cdot x_i^{(l + 1)} \right\rfloor} \right)
\end{equation}
along the $i$-th path according to Eqn.~\ref{Eqn:Sobol}.
Similar to generating the paths by a pseudo-random number generator,
the connectivity of the network does not need to be stored explicitly, because
the components of the Sobol' sequence in Eqn.~\ref{Eqn:Sobol} can be
computed on the fly. When the numbers of neurons in the input or output layer are not
powers of two, one may choose to just fully connect these layers
with their corresponding hidden layers.

The example in Fig.~\ref{Fig:ANNprogressive} demonstrates one advantage of encoding
the network topology by a low discrepancy sequence. As the
permutations are progressive, it is straightforward to add another
power of 2 connections. Enumerating the network topology
from sparse to fully connected becomes natural.

Fig.~\ref{Fig:ANNarchitectures} shows an example of
a sparse classifier network generated by the Sobol' sequence. A high dimensional input vector
on the left is condensed to a vector of classes on the right.
Each layer has a power of 2 neurons and each neuron in a layer
has the exact same constant number of connections.
The next example is an autoencoder structure that is often used
for filtering signals. Input and output layer are of the
same dimension.

Creating a sparse network with non-negative weights like in Sect.~\ref{Sec:NonNegative}
is as simple as selecting the first half of the paths to have non-negative
weights and the second half to have non-positive weights. A second option is
to dedicate one dimension of the Sobol' sequence to determine whether the weights
of a path shall be non-negative or non-positive just by checking whether
the component is smaller than $\frac{1}{2}$ or not. More details on
partitioning one low discrepancy sequence into many are found
in \cite{ParQMC}. If the number of paths is a power of 2, partitioning a network
generated by the Sobol' sequence into supporting and
inhibiting network as described will result in a zero sum of weights
per neuron if neurons in a layer have constant valence. This nicely complements
the normalized initialization (see Sect.~\ref{Sec:Init} and
Sect.~\ref{Sec:Normalize}) and typically is not guaranteed when using a
pseudo-random number generator to generate the paths.

When the number of paths exceeds the product of the number of neurons
in two successive layers, edges will be selected more than once.
Even before reaching that bound and although the components of the
Sobol' sequence create progressive permutations, it may happen that
multiple edges as defined by Eqn.~\ref{Eqn:Link} coincide.
While coalescing edges are not a problem for the algorithm in Fig.~\ref{Fig:Code},
having multiple weights
associated to one edge is redundant and may reduce the capacity of the network.
This reason for this issue has been known for a long time in the domain of quasi-Monte Carlo 
methods and especially from the Sobol' sequence, where low dimensional projections
may expose very regular correlation patterns between the dimensions.
To improve on the issue, low discrepancy sequences have been scrambled \cite{Owen:1995} and optimized \cite{JK03,Myths04}.
We have been successful by simply omitting
the dimensions of the Sobol' sequence whose generator matrices
result in coalescing edges, which can be interpreted as a permutation
of the sequence of the Sobol' generator matrices. Using the cascaded construction
of Sobol' points \cite{Cascaded} that forces each successive pair of
dimensions to form a $(0,m,2)$-net, coalescing
edges are avoided by construction. The respective numerical results
are presented in Sect.~\ref{Sec:Paths_in_CNN}.

\subsection{Hardware Considerations} \label{Sec:CollisionFree}

Representing sparse networks by paths, the algorithm in Fig.~\ref{Fig:Code}
linearly streams the weights
from memory. Such an access pattern perfectly matches the parallel loading
of contiguous blocks of weights in one cache line by the pre-fetcher as it is
common for current processor hardware.

Similar to \cite{dey2017edgeProcessing} and \cite[Fig.~4]{SparseNNHardware}, the permutations
generated by the Sobol' sequence guarantee streaming weights in
contiguous blocks of size of a power of 2 free of memory bank conflicts.
For the same reason, weights can be routed without collisions
through a crossbar switch inside the processor. Both advantages cannot
be guaranteed when creating paths by pseudo-random number generators.

Determining the permutations generated by $\Phi_2$ in hardware amounts
to bit reversal, which is straightforward to hardwire. Implementing the permutations
generated by Eqn.~\ref{Eqn:Sobol} in hardware requires to unroll
the loop over the bit-parallel XOR operations (see the algorithm in Sect.~\ref{Sec:LD}).
This results in a tiny circuit with a matrix of flip-flops to hold
the generator matrix $C_j$. Replicating parts of the circuit for all
numbers representable by the $m$ least significant bits allows
one to create $2^m$ values of the permutation in parallel.

For backpropagation, we can take advantage of the fact the Sobol' sequence
is invertible. Computing the inverse of Eqn.~\ref{Eqn:Sobol} just requires to
determine $C_j^{-1}$.
Propagating errors back through the network, the memory access
remains contiguous when enumerating the array of weights backwards.

\section{Numerical Results} \label{Sec:Results}

We perform numerical experiments to evaluate the accuracy
of neural networks represented by paths generated by the Sobol'
low discrepancy sequence. We take a look at classification tasks using
classic multilayer perceptrons (MLP) as illustrated in Fig.~\ref{Fig:ANN}, convolutional
neural networks, the set of hyperparameters, and the initialization of sparse neural networks.

\subsection{Performance of Sparse Neural Networks represented by Paths}

Using the algorithm in Fig.~\ref{Fig:Code}, we demonstrate the linear complexity of sparse neural networks represented by paths
as compared to fully connected networks represented by matrices.
While it is possible to compress sparse matrices, there is an additional cost for decompression that needs to be amortized.
In contrast, representing the neural network by paths, a linear speedup results for any desired sparsity.
Fig.~\ref{Fig:sparseANN-by-paths-speedup} compares the relative runtimes for 1 epoch for a network of four layers with 256 neurons per layer, varying the number of paths,
and compares them to the sparsity and accuracy. The experiments were run on a single core of an AMD 7 5800X CPU.

\begin{figure}
  \centering
  \begin{tikzpicture}
    \begin{axis}[
      legend cell align=left,
      ylabel style={align=center},
      xlabel style={align=center}, xlabel=Number of paths per neuron, 
      xtick={4, 8, 16, 32, 64, 128, 256},
      legend pos=south east, 
      legend style={draw=none}, 
      xmin=0, xmax=256,
      ymin=0, ymax=100, 
      height=4.5cm, width=\linewidth,
    ] 
      \addplot[thick,color=HighlightColor,mark=*] coordinates {
        (4,	89.09) (8,	95.76) (16, 96.17) (32, 96.57) (64, 96.89) (128, 97.59) (256, 97.64)};\label{plot_acc}
      \addlegendentry{Accuracy in \%}
      \addplot[thick,color=HighlightColor2,mark=*] coordinates {
        (4,	1.56) (8,	3.125) (16, 6.25) (32, 	12.5) (64, 25.00) (128, 50.0) (256, 100.0)};
      \addlegendentry{100\% - Sparsity}
      \addplot[thick,color=HighlightColor3,mark=*] coordinates {
        (4,	2.19) (8,	4.78) (16, 8.89) (32, 	16.74) (64, 28.74) (128, 46.87) (256, 100.0)};
      \addlegendentry{Relative runtime in \%}
  \end{axis}
  \end{tikzpicture}
  \caption{Accuracy, sparsity and runtime relative to the fully connected neural network for image classification on MNIST for sparse networks represented by paths with 4 hidden layers of 256 neurons.}
  \label{Fig:sparseANN-by-paths-speedup}
\end{figure}

\subsection{Training Sparse from Scratch} \label{Sec:Classifier}

For the simple examples of recognizing digits in tiny images (see
Fig.~\ref{Fig:sparseANN-by-paths-speedup}), training a sparse-from-scratch artificial
neural network reveals that only a tiny fraction of paths as compared to the number of
connections within the fully connected network is required to come very close to
the test accuracy of the fully connected variant.

Comparing paths generated by pseudo-random numbers to paths generated by a low
discrepancy sequence does not show a big difference in accuracy compared to random walks. However, as stated in
Sect.~\ref{Sec:QRN}, using the Sobol' low discrepancy sequence allows for routing
without bank conflicts and avoids duplicate weights. This guarantee is also a big advantage over pseudo-random
number generated access patterns when considering a hardware implementation
\cite{dey2017interleaver,SparseNNHardware}.

\begin{figure}
  \centering
  \begin{tikzpicture}
    \begin{axis}[ylabel=Test Accuracy (\%), xlabel=Number of paths, 
      legend cell align=left,
      legend pos=south east, legend style={draw=none}, 
        xmin=0, xmax=4200,
        ymin=0, ymax=100,
        height=4.5cm, width=\linewidth,
        xtick = {64, 256, 1024, 2048, 4096}] 
        \addplot[thick,color=HighlightColor2,mark=*] coordinates { (4200, 87.27)};
        \addlegendentry{fully connected}
        \addplot[thick,color=HighlightColor4,mark=*] coordinates {
           (64,	64.16) (256, 	76.52) (1024,	83.71) (2048,  85.43) (4096, 86.48)};
        \addlegendentry{pseudo-random paths}
        \addplot[thick,color=HighlightColor3,mark=*] coordinates {
          (64, 63.05)  (256,	78.21) (1024, 	83.81) (2048,	83.99) (4096, 85.98)};
        \addlegendentry{quasi-random paths}
        \addplot[thick,color=HighlightColor5,mark=*] coordinates {
          (64, 63.05) (256, 78.21) (1024,	84.88) (2048,  86.07) (4096, 87.27)};
        \addlegendentry{quasi-random paths, no coalescing edges}
        \addplot[thick,dashed,color=HighlightColor,mark=o,mark options={solid}] coordinates {
          (64,	63.69) (256, 	78.00) (1024,	85.03) (2048,  86.58) (4096, 87.27)};
        \addlegendentry{quasi-random paths, Cascaded Sobol'}
    \end{axis}
  \end{tikzpicture}
  \caption{Test accuracy of a convolutional neural network (CNN) represented by paths
    and trained sparse from scratch compared to the fully connected
    counterpart. The task is recognizing 10 classes of objects in $32 \times 32$
    pixel images (CIFAR-10). Around 1024 paths the accuracy reaches a plateau very close to the maximum accuracy, advocating sparse networks.}
  \label{Fig:CNNsparsefromscratchAccuracy}
\end{figure}

\begin{figure}
  \centering
  \begin{tikzpicture}
    \begin{axis}[
      ylabel=Number of unique edges, 
      xlabel=Number of paths,
      legend cell align=left,
      legend pos=south east, legend style={draw=none, cells={align=left}, fill=none, font=\scriptsize}, legend columns=2,
      xmin=0, ymin=0, ymax=75000, xmax=4200, height=4.5cm, width=\linewidth,
        xtick = {64, 256, 1024, 2048, 4096}]
        	\addlegendimage{empty legend}
        \addlegendentry{}
	\addplot[thick,color=HighlightColor2,mark=*] coordinates { (4200, 70618)};
        \addlegendentry{fully connected}
        \addplot[thick,color=HighlightColor4,mark=*] coordinates {
          (16,	1001) (64,	2954) (256, 	9318) (1024, 	26732) (2048, 40232) (4096, 54282)};
        \addlegendentry{random paths}
        \addplot[thick,color=HighlightColor3,mark=*] coordinates {
          (16,	1034) (64,	3226) (256, 	10330) (1024, 	29128) (2048, 29146) (4096, 52168)};
        \addlegendentry{quasi-random paths}
        \addplot[thick,color=HighlightColor5,mark=*] coordinates {
          (16,	1034) (64,	3226) (256, 	10330) (1024, 	33754) (2048, 52186) (4096, 70618)};
        \addlegendentry{quasi-random paths, \\no coalescing edges}
        \addplot[thick,dashed,color=HighlightColor,mark=o,mark options={solid}] coordinates {
          (16,	1034) (64,	3226) (256, 	10330) (1024, 	33754) (2048, 52186) (4096, 70618)};
        \addlegendentry{quasi-random paths,\\ Cascaded Sobol'}
    \end{axis}
  \end{tikzpicture}
    \caption{Sparse networks created with random paths may have some parts of the paths overlap, creating coalescing edges.
    For the Sobol' sequence it may occur that certain combinations of dimensions result in coalescing edges as well (blue line).
    This can be resolved by simply omitting the dimensions of concern (purple line). Cascaded Sobol' points avoid coalescing edges by construction.
    Note that for 1024 paths the accuracy (see Fig.~\ref{Fig:CNNsparsefromscratchAccuracy}) already has reached a
  plateau.}
  \label{Fig:CNNsparsefromscratchParameters}
\end{figure}

\subsection{Random and Quasi-Random Paths in CNNs} \label{Sec:Paths_in_CNN}

Similar to the fully connected neural networks, convolutional neural networks
(CNNs) represented by paths
(see Sect.~\ref{Sec:CNN}) can be trained sparse from scratch.
For the numerical experiments, we use the CIFAR-10 image recognition data set.
Our CNN has 5 convolutional layers with a number of channels 16,
32, 32, 64, 64, respectively, followed by one fully connected layer with a softmax activation function to produce 10 output
features each identifying one of the dataset classes. Every convolutional layer
is followed by a Batch Norm layer \cite{ioffe2015batch} and a ReLU activation function
\cite{glorot2011deep}. Training is done using stochastic gradient descent (SGD) with a momentum
of 0.9 and weight decay of 0.0001, for 182 epochs. The learning rate starts at 0.1, and is decreased by factor 10 at epochs
91 and 136.
We normalize the input images using the mean and standard deviation over the
training set, and apply additional augmentation in the form of random horizontal
flips as well as a $32\times32$ crop after padding the input image 4 pixels on every side.

Fig.~\ref{Fig:CNNsparsefromscratchAccuracy} shows the accuracy of sparse from scratch
CNNs compared to a fully connected CNN, and Fig.~\ref{Fig:CNNsparsefromscratchParameters} shows the corresponding number of
non-zero parameters. We observe a sharp increase in accuracy initially and then a slower convergence
towards the accuracy of the fully connected network.  Importantly, 
the graphs show that an accuracy close to the fully connected network can be
reached with far fewer weights by sparse networks.
The figure also shows that sampling paths randomly or
quasi-randomly using the Sobol' sequence performs very
similarly in terms of accuracy. However, as remarked in
Sect.~\ref{Sec:QRN}, there are large potential hardware advantages when using
quasi-random methods.

\subsection{Relation of Number of Paths, Layer Width, and Accuracy}

It has been observed that wider but sparse networks have higher
representational capacity than narrower dense networks \cite{gray2017gpu,child2019generating}. 
Selecting the convolutional neural network as in the previous section, we try to verify 
this empirically in Table~\ref{tab:CNNwidesparseVSnarrowdense}, where
we investigate how the accuracy changes when we scale the number of neurons per layer, 
i.e. the network width, but keep the number of weights constant.
This can be done by increasing the sparsity as the width increases.
In a fully connected network the number of weights increases quadratically with width, whereas 
for networks defined by paths the number of weights is determined by the number of paths.
The experiment shows that some sparse models can achieve better accuracy than the fully connected model, 
but accuracy starts degrading when the networks get too sparse.

\begin{table}
  \centering\small
  \begin{tabular}{lcccc}
    Width multiplier & Number of paths & Sparsity  & Test accuracy   & Test loss \\ \toprule
    1.0             & fully connected & 0\%         &   87.27\%       &   0.384     \\
    1.25            & 4150            & 35.51\%     &   87.60\%       &   0.387     \\
    1.5             & 3050            & 55.12\%     &   87.12\%       &   0.392     \\
    2.0             & 2420            & 74.64\%     &   86.93\%       &   0.399     \\
    4.0             & 1950            & 93.59\%     &   86.68\%       &   0.408     \\
    8.0             & 1800            & 98.37\%     &   84.86\%       &   0.448     \\\bottomrule
  \end{tabular}
  \caption{Comparing fully connected to wider, sparser networks created
  by random paths. The number of paths is chosen such that all networks have around 70400 weights like the fully connected
  network. All networks are close, but highest accuracy is achieved for width multiplier 1.25.}
  \label{tab:CNNwidesparseVSnarrowdense}
\end{table}

\begin{table}[t]
  \centering
  \begin{tabular}{llr}
    CNN & Initialization method & Test accuracy\\ \toprule
    Dense   & Uniformly random        & 87.27\% \\
       & Constant, positive           & 10.00\%\\
       & Constant, alternating sign   & 10.00\%  \\
       & Constant, random sign        & 86.88\% \\
       & Constant, random sign, 90\% sparse & 87.39\% \\
    \midrule
    Sparse  & Uniformly random       & 83.71\% \\
      & Constant, positive           & 82.15\% \\
      & Constant, alternating sign   & 83.70\% \\
      & Constant, random sign        & 83.40\% \\
      & Constant, sign along path    & 83.75\% \\
    \midrule
    Sparse, signs fixed   & Constant, alternating sign & 80.77\% \\
    (train only magnitude) & Constant with constant sign along path   & 77.61\% \\
    \bottomrule
  \end{tabular}
  \caption{Comparison of initializing weights uniformly random, constant positive, constant
  with half the initial values negative, or constant but positive for odd neuron
  indices and negative otherwise ("alternating"). The task is image classification
  (CIFAR-10). The sparse convolutional neural network (CNN) is created by tracing 1024
  random paths. These sparse neural networks have 26.4K weights as compared to
  70.6K for the dense net. ``Signs non-trainable'' means that signs are kept fixed after initialization,
  while training only the weight magnitudes.}
  \label{tab:CNNinitialization}
\end{table}

\subsection{Initialization of Structurally Sparse Networks}

Verifying the analysis from Sect.~\ref{Sec:Init} and Sect.~\ref{Sec:NonNegative},
we compare different initialization strategies and show the results in Table~\ref{tab:CNNinitialization}.

For the fully connected networks, setting all weights to the same value prevents
learning as expected. Similarly, using the same magnitude but setting
the sign positive for weights with even index and negative otherwise
("alternating") makes the network too regular to learn. However, we see that simply
choosing a random sign can result in very high accuracy, provided the constant $w_{\text{init}}$ is chosen
carefully as described in Sect.~\ref{Sec:Init}.
We also find that initializing the weights sparse doesn't hurt accuracy, and in fact seems to slightly improve it. 
With 90\% sparsity, it may happen that a weight slice (see Sect.~\ref{Sec:CNN}) is completely zero at initialization.
This is not a problem as long as there are non-zero values in the other filters belonging to that neuron.

For the sparse networks, we can make use of the paths for initialization, setting
weights belonging to a path with an even index positive and negative otherwise.
This can be done only for initialization
or permanently in order to save one bit of storage per weight. 
The last 2 rows of Table~\ref{tab:CNNinitialization} show what happens if the signs
of the weights are fixed and we only train the weight magnitudes. The
signs can be stored or generated dynamically as described in
Sect.~\ref{Sec:NonNegative}. Training only weight magnitudes, 
while initializing all weights with the same constant, the networks still
reach accuracy within 3\% of the fully randomly initialized network.

Care needs to be taken when choosing to fix the sign for all weights along a path in a CNN.
As a path touches not a single weight but a $w \times h$ depth slice (see Sect.~\ref{Sec:CNN}), enforcing the same sign
for all these weights prevents the network from learning many types of features like
edges. Still, such a sparse network using random paths is able to reach 80\% accuracy.
Note that this is not an issue for the common case of $1 \times 1$-convolutions, where $w = h = 1$.

The sparse networks are far more robust to the initialization and do not fail in
any of the cases even with all weights in a layer set to the same positive constant.
Using a deterministic low discrepancy sequence to enumerate the paths and
to determine the signs of the weights allows for deterministic initialization
and hence brings us one step closer to completely deterministic training.

\section{Conclusion}

Encoding the network topology by a deterministic low discrepancy sequence
brings together quasi-Monte Carlo methods and artificial neural networks.
The resulting artificial neural networks may be trained much more
efficiently, because they are structurally sparse from scratch.
In addition they allow for deterministic initialization. As shown for the example
of the Sobol' sequence, the resulting memory access and connection patterns are
especially amenable to a hardware implementation, because they guarantee
collision-free routing and constant valences across the neural units.

In future work, we will extend the investigations of quasi-Monte
Carlo methods applied to other types of neural networks.
Especially in the domain of speech recognition,
preliminary experiments are very promising.
Furthermore, we like to look at more low-discrepancy sequences and at 
growing neural networks during training by progressively
sampling more paths as generated by the low discrepancy sequence.

\begin{acknowledgement}
The first author is very thankful to C\'edric Villani for a discussion
on structure to be discovered in neural networks
during the AI for Good Global Summit 2019 in Geneva.
The authors like to thank Jeff Pool, Nikolaus Binder, and David Luebke for profound discussions and Noah Gamboa, who helped with
early experiments on sparse artificial neural networks. This work has
been partially funded by the Federal Ministry of Education and Research (BMBF, Germany)
in the project Open Testbed Berlin - 5G and Beyond - OTB-5G+ (F\"orderkennzeichen 16KIS0980).
\end{acknowledgement}

\end{document}